\documentclass[conference]{IEEEtran}
\IEEEoverridecommandlockouts
\usepackage{amsmath,amssymb,amsfonts}
\usepackage{algorithmic}
\usepackage{graphicx}
\usepackage{textcomp}
\usepackage{xcolor}
\usepackage{booktabs}
\usepackage{lipsum}
\usepackage{fancyhdr}

\def\BibTeX{{\rm B\kern-.05em{\sc i\kern-.025em b}\kern-.08em
    T\kern-.1667em\lower.7ex\hbox{E}\kern-.125emX}}

\begin{document}

\title{Maximum Entropy Hindsight Experience Replay\\

\thanks{The authors would like to thank Megan Emmons and Srideep Musuvathy for helpful discussions.
This work was supported by the Laboratory Directed Research and Development program at Sandia National Laboratories, a multimission laboratory managed and operated by National Technology and Engineering Solutions of Sandia LLC, a wholly owned subsidiary of Honeywell International Inc. for the U.S. Department of Energy's National Nuclear Security Administration under contract DE-NA0003525. The authors own all right, title and interest in and to the article and is solely responsible for its contents. The United States Government retains and the publisher, by accepting the article for publication, acknowledges that the United States Government retains a non-exclusive, paid-up, irrevocable, world-wide license to publish or reproduce the published form of this article or allow others to do so, for United States Government purposes. The DOE will provide public access to these results of federally sponsored research in accordance with the DOE Public Access Plan https://www.energy.gov/downloads/doe-public-access-plan.  This paper describes objective technical results and analysis. Any subjective views or opinions that might be expressed in the paper do not necessarily represent the views of the U.S. Department of Energy or the United States Government. $\dagger$ dccrowd@sandia.gov. SAND2024-14947O
}
}

\author{
\IEEEauthorblockN{Douglas C. Crowder$\dagger$}
\IEEEauthorblockA{\textit{Cognitive and Emerging Computing} \\
\textit{Sandia National Laboratories}\\
Albuquerque, NM} \\

\IEEEauthorblockN{Darrien M. McKenzie}
\IEEEauthorblockA{\textit{Autonomy for Hypersonics} \\
\textit{Sandia National Laboratories}\\
Albuquerque, NM}

\and
\IEEEauthorblockN{Matthew L. Trappett}
\IEEEauthorblockA{\textit{Autonomous Sensing and Control} \\
\textit{Sandia National Laboratories}\\
Albuquerque, NM} \\

\IEEEauthorblockN{Frances S. Chance}
\IEEEauthorblockA{\textit{Cognitive and Emerging Computing} \\
\textit{Sandia National Laboratories}\\
Albuquerque, NM}
}

\maketitle

\begin{abstract}
Hindsight experience replay (HER) is well-known to accelerate goal-based reinforcement learning (RL).  While HER is generally applied to off-policy RL algorithms, we previously showed that HER can also accelerate on-policy algorithms, such as proximal policy optimization (PPO), for goal-based Predator-Prey environments.  Here, we show that we can improve the previous PPO-HER algorithm by selectively applying HER in a principled manner.
\end{abstract}
\section{Introduction}

Reinforcement learning (RL) is a biologically-inspired artificial intelligence (AI) algorithm that allows agents to learn by interacting with the environment \cite{sutton2018reinforcement}.  In RL, an agent observes an environment and chooses actions that get applied to the environment, causing the environment to transition to a new state and emit a reward.  The job of the RL agent is to maximize the time-discounted expected sum of future rewards \cite{sutton2018reinforcement}.  RL has already allowed machines to achieve super-human performance in many diverse tasks \cite{mnih2013playing, mnih2015human, lillicrap2015continuous}.

One of the major drawbacks of RL is the poor sample efficiency; RL algorithms often require many samples to learn tasks that are simple for humans to learn.  When interacting with the real-world, it can be expensive to generate samples because RL agents can choose bad actions that lead to real-world consequences.  Similarly, generation of samples in high-fidelity simulations can also be quite expensive because of the time required to perform the necessary calculations.  Thus, one of the main goals of RL is to decrease the number of samples required to become proficient at a task.

A common method for increasing the sample efficiency of RL is to use reward shaping, where rewards are chosen to make the task easier to learn.  However, if rewards are shaped inappropriately, shaped rewards can cause RL agents to converge to sub-optimal policies \cite{ng1999policy}, and choosing appropriately-shaped rewards can be non-trivial.  Thus, hindsight experience replay (HER) was developed to make goal-based RL more sample-efficient while avoiding the use of shaped rewards \cite{andrychowicz2017hindsight}.  HER involves changing the goal of an RL agent after an episode has been completed to be some state achieved during the episode.  This allows the RL agent to learn how to complete trajectories, even if the completed trajectories were not originally desired.  When using function approximators, such as neural networks, this can allow RL agents to interpolate between learned trajectories, enabling them achieve the desired goals.

Typically, HER is only applied to off-policy algorithms such as DQN \cite{mnih2013playing}, DDPG \cite{lillicrap2015continuous}, TD3 \cite{fujimoto2018addressing}, and SAC \cite{haarnoja2018soft}.  Compared to on-policy algorithms, such as PPO \cite{schulman2017proximal}, off-policy algorithms tend to be more sample efficient, but less clock-time efficient, particularly when environments are computationally-efficient to evaluate.  Compared to off-policy algorithms, PPO is frequently more stable.  However, on-policy algorithms make the assumption that episodes within the training buffer were generated by the current RL policy.  This assumption is violated when HER modifies the goal \textit{post hoc}.  However, we showed in previous work that, for certain Predator-Prey environments, HER can be applied to PPO, despite violating the assumptions of PPO \cite{crowder2024ppoher}.  In that previous work, PPO-HER generally outperformed PPO, SAC, and SAC-HER.  However, for some of the more challenging environments, PPO converged to a local optima.

The purpose of this work is to modify HER to avoid those local optima.  Specifically, we use information theory to select transitions from the rollout buffer that maximize the information available to HER.  We show that, using this so-called ``maximum entropy HER'' (MEHER), we can avoid local optima.

\begin{figure*}
    \centering
    \includegraphics[width=\textwidth]{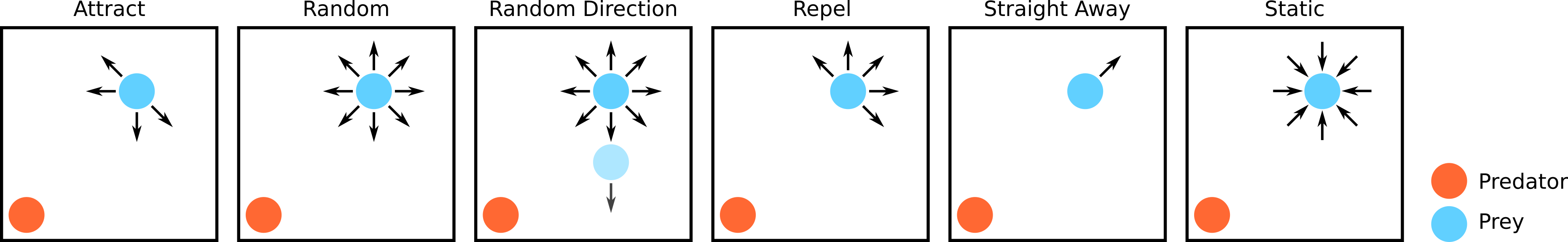}
    \caption{Predator-prey environments.}
    \label{fig:task}
\end{figure*}

\section{Methods}

\subsection{Task}

We used a series of 3-dimensional Predator-Prey tasks, as illustrated in Figure \ref{fig:task} and described previously \cite{crowder2024ppoher}.  Briefly, a Predator agent learns via reinforcement learning to intercept a Prey agent that is executing one of a few fixed policies.  The Predator agent can move at twice the maximum speed of the Prey.  At the initial stage, the Prey and Predator either spawn at random positions, sampled from a uniform distribution across all possible positions (Spawn Random) or spawn at fixed locations (Spawn Apart).  Interception can be quite easy or quite difficult, depending on the Prey policy.  For instance, the Attract policy causes the Prey to move in the general direction of the Predator (a fairly easy task) whereas the Straight Away policy causes the Prey agent to move in a straight line away from the Predator, which is an optimal policy for evading the Predator, given that both agents are equally maneuverable.  The Straight Away task is very difficult because RL agent kinematics resemble random walks at the beginning of training, which will result in very few accidental interceptions of the Prey.  Episodes terminate when the Predator intercepts the Prey or after a fixed number of timesteps.  The environment emits a reward of 1 if the Predator intercepts the Prey, a reward of -1 if Predator fails to intercept the Prey before the end of the episode, and a reward of 0 for all non-terminal timesteps.

\subsection{Reinforcement Learning}
We used a high-quality implementation of PPO \cite{stable-baselines3} with a custom HER implementation, as describe previously \cite{crowder2024ppoher}.  We used the ``final'' HER method.  We refer to the combination of PPO and HER as PPO-HER.  All simulations conditions were repeated 12-16 times.

\subsection{Terminology}
In this work, we evaluate the performance of RL agents using a few metrics, including the ``success rate.''  During training, we manipulate the data in the buffer to achieve a given ratio of successes, which we call the ``S-ratio.''  Success rate is a dependent variables, whereas S-ratio is an independent variable.

\subsection{Maximum Entropy HER}
To optimize the performance of PPO-HER, we make use of the principle of entropy from information theory.  Information theory quantifies the amount of information in a random signal $X$ by using a quantity called ``entropy'' $H(X)$, which is defined as:

\begin{align}
    H(X) := -\sum_{x\in\chi} p(x) \text{log} p(x)
\end{align}

where samples of $X$, $x$, are taken from the set $\chi$.  Analysis of this equation yields the fact that entropy (and therefore information) in a signal is maximized when $p(x_i) = p(x_j) \forall x_i, x_j \in \chi$.  In other words, entropy (and information) in a signal are maximized when elements of the set $\chi$ are drawn from a uniform distribution.  Minimum entropy is achieved when the probability distribution is a delta function, meaning that the probability of observing 1 symbol in the signal is 1, and the probabilities of observing all other signals is 0.  These results hold when $X$ is drawn from a continuous distribution instead of a discrete distribution.  For a discrete binomial distribution composed of 2 symbols, $x_1$ and $x_2$, the plot of entropy as a function of $p(x_1)$ is a parabola that is maximized at $p(x_1)=p(x_2)=0.5$.

RL agents use several signals to learn.  These signals include vector observations of the environment states, vector actions taken by the the agent, and a scalar reward that quantifies the performance of the agent.  These signals are used to train the neural networks that underlie deep RL algorithms, such as PPO.  We are interested in maximizing the amount of information contained in the signals used to train RL agents - thus, we want to maximize the entropy in these signals.

In PPO, actions are drawn from Gaussian distributions, where the RL agent specifies the mean and standard deviation of the Gaussian.  If the standard deviation is sufficiently large, and if the actions are bounded, this Gaussian distribution may resemble a uniform distribution with 2 delta functions at either extreme of the bounds (caused by actions being clipped).  While it might seem interesting to replace the Gaussian distribution with a uniform distribution, this is impractical since training gradients are calculated as the ratio of the probabilities of actions under the current and the previous action distributions.  If we were to switch to a uniform distribution, these gradients would become quite unstable since most points under a uniform distribution have 0 probability.  Thus, calculating the probability ratio between 2 uniform distributions would result in many gradients that are 0 or undefined.  Given this fact, it was not immediately apparent how to apply information theory to the action signal.  So, we continued by specifying Gaussian distributions with initial standard deviations of 1.  Actions were bounded to the range $[-1, 1]$.

The observation vector $o$ is composed of 2 sub-vectors: a sub-vector of Predator and Prey kinematics $s$, and a sub-vector of the Predator goal (the Prey position) $g$, yielding $o = [s, g]$.  HER works by replacing $g$ with some state achieved during the episode, $g'$.  At most time steps, for most of the environment conditions, we cannot directly control $s$ for the purposes of increasing the signal entropy because $s$ is determined by the current Predator and Prey policies.  However, for the Random Spawn conditions, the initial positions of the Predator and Prey were drawn from a uniform (high-entropy) distribution.  Additionally, for the Random Prey policy, the Prey moved in a random direction at each time step with a random magnitude.  Therefore, while there are limits to the amount of entropy that we can injected into the observation signal, some environment conditions existed that had higher-than-average entropy.

Although we could not easily inject entropy into the observation and action signals, we were able to inject entropy into the reward signal.  At each time step, the Predator received a reward from the set $\{-1, 0, 1\}$.  If we consider the episode returns (the sum of all rewards received over each episode), each return was drawn from the set $\{-1, 1\}$ since all episodes were either failures or successes, and each episode terminated upon success or failure.  We resampled the data to artificially increase or decrease the amount of entropy in the return signal.  To increase the number of successes, we used the ``finall'' HER method to resample the goal \cite{andrychowicz2017hindsight}.  To increase the number of failures, we resampled the goal from a uniform distribution across the environment positions.  After resampling the goals, we could filter the data in the training buffer to perfectly achieve the desired ratio of successes to failures.  Within the filtered training buffer, the fraction of successful transitions is refered to as the S-ratio.

According to the information theory described above, we hypothesized that we could improve the performance of RL agents, as well as the learning rate, by achieving an S-ratio of 0.5.  This hypothesis is supported by the observation that, for our Predator-Prey environment, successful training of RL agents results in a sigmoidal learning curve (a plot of achieved reward as a function of time).  The highest derivative of sigmoidal curves is at the center of the curve (on the y-axis, which represents rewards).  Thus, the highest learning rate is associated with an S-ratio of 0.5, achieved naturally over the course of RL.  To test this hypothesis, we varied the success probabilities within the range $[0, 1]$.

This information theory hypothesis predicts that, if we plot maximum rewards as a function of the S-ratio, this will results in a parabola with a maximum at an S-ratio of 0.5.  For very simple tasks, a wide variety of success probabilities will result in achieving the maximum reward.  Thus, to investigate the effects of S-ratio on performance we use a custom metric $M_c$ for each condition $c$ in the set of conditions $C$ based on the normalized maximum median success rate $R_c$ and the normalized time to learn $T_c$, defined as the number of steps required to achieve a success rate of $0.95 \times R_c$:

\begin{align}
    M_c = 
            \frac{R_c}{\text{max}(R_C)}
            \times 
            \left(1 - \frac{T_c}{\text{max}(T_C)}\right)
    \label{eq:mc}
\end{align}

This metric is high when performance is high and time to learn is low.  If a given S-ratio results in the maximal reward and the maximal learning rate, $M_c = 1$.  If a given S-ratio results in the minimal reward and the minimal learning rate, $M_c = 0$.  The $M_c$ metric suffers from a bias that agents that do not learn (meaning that they are always within approximately 95\% of the maximum median episode return that they ever achieve) appear to learn more quickly.  Thus, rather than considering the absolute magnitude of $M_c$, one should only compare the relative magnitudes of $M_c$ within a given environmental condition.

\subsection{Targeted Maximum Entropy HER}
Our initial experiments revealed that S-ratios greater than 0.5 frequently performed better than S-ratios less than 0.5, suggesting that successes are more informative than failures.  Note that information theory suggests symmetry about 0.5 such that an S-ratio of 0.2 should result in the same performance as an S-ratio of 0.8.  To investigate this phenomenon further, we attempted to make failures more informative.  We reasoned that near misses are more informative than far misses because there are relatively few ways to miss by a little and relatively many ways to miss by a lot.  Thus, for the ``targeted'' MEHER condition, when resampling goals for failures, instead of drawing new goals from a uniform distribution across the environment, we chose goals that were 1.1 times the Predator-Prey interception distance away from the prey.

\subsection{PPO-HER-2-PPO}
Previous work \cite{crowder2024ppoher} revealed that PPO-HER adds computational overhead, compared to PPO, because PPO-HER requires resampling goals.  For simple Predator-Prey environments, which are quick to evaluate, this overhead slowed learning by as much as a factor of 2.  To determine if we could decrease the computational overhead of PPO-HER while improving performance, we stopped using HER once the agent was successful in 50\% of the episodes.  We refer to this as PPO-HER-2-PPO.

\begin{figure*}
    \centering
    \includegraphics[width=\textwidth]{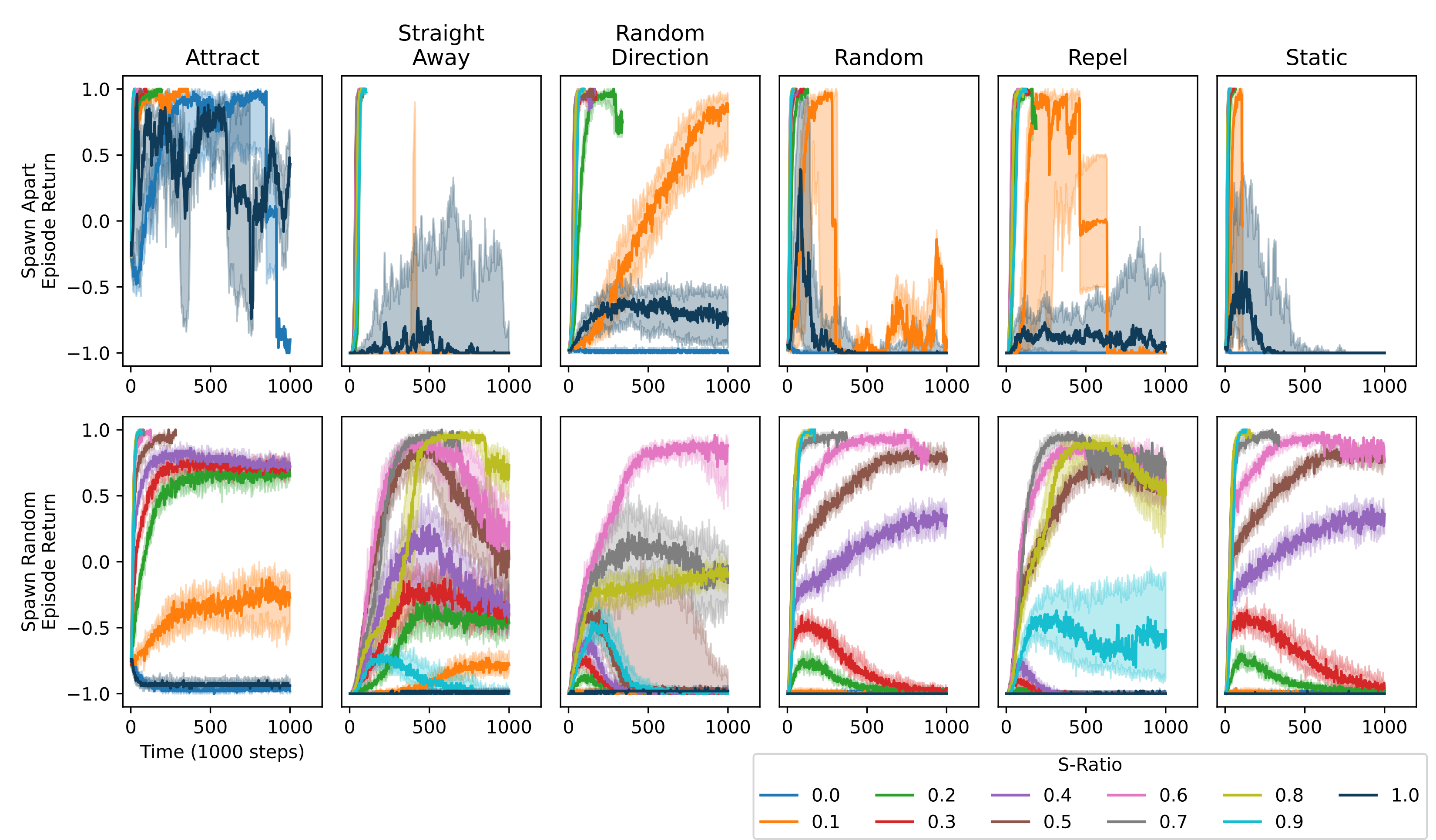}
    \caption{Learning curves for agents trained using S-ratios in the range $[0, 1]$.  Across conditions, S-ratios in the range $[0.5, 0.7]$ tend to perform best, with a slight bias towards higher S-ratios.  Bold lines and shaded regions represent medians and interquartile ranges, respectively.}
    \label{fig:meher}
\end{figure*}

\begin{figure*}
    \centering
    \includegraphics[width=\textwidth]{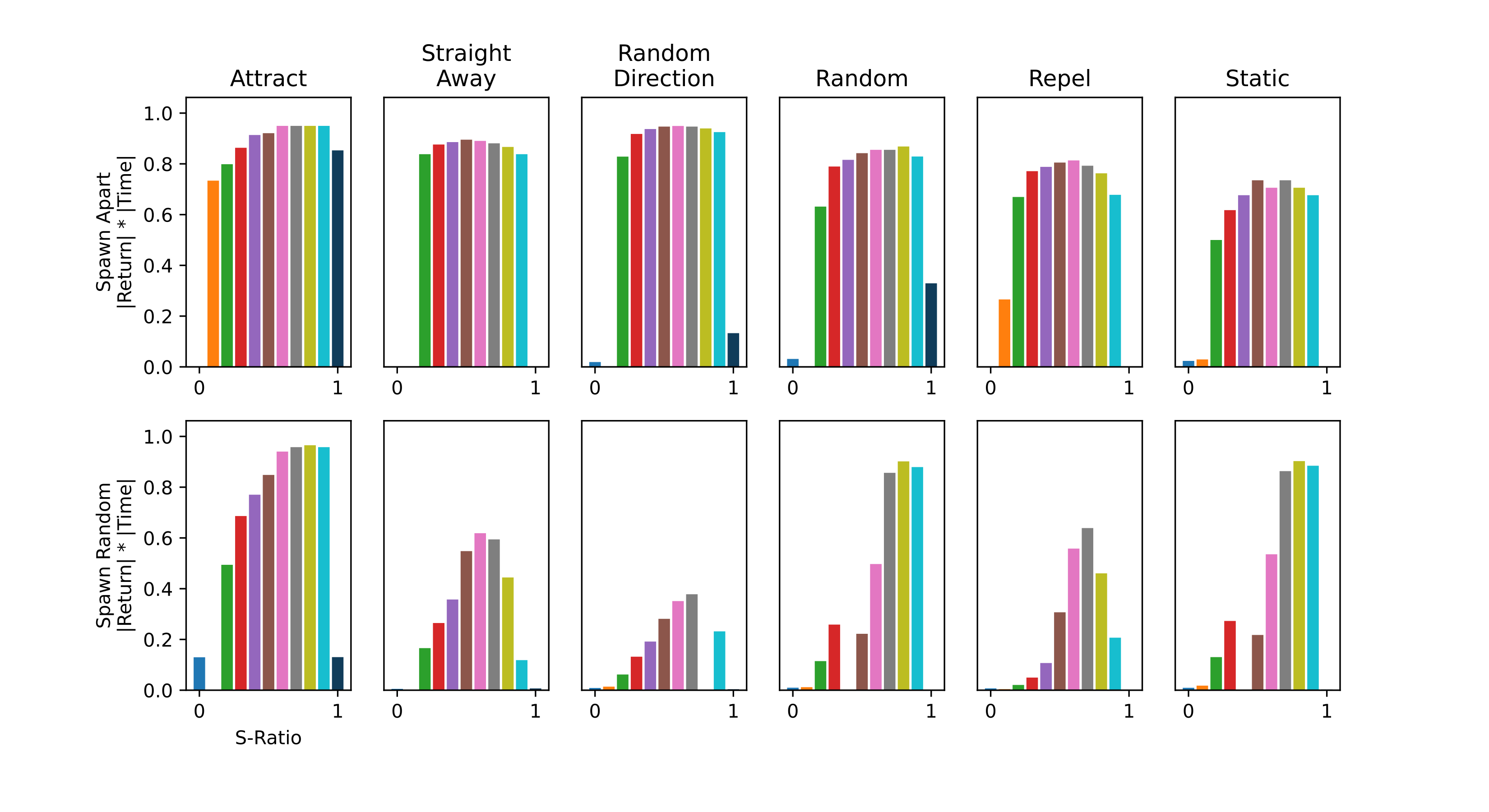}
    \caption{$M_c$ metrics for S-ratios in the range $[0, 1]$.  The shape of the bar plots, rather than the magnitudes of the columns, should be interpreted.  Information theory predicts a parabolic shape to the bar plot, maximized at an S-ratio of 0.5, with $M_c$ values of 0 for S-ratios of 0 and 1.  The actual shapes of the bar plots differ somewhat from parabolas, and higher S-ratios tend to perform better than predicted.}
    \label{fig:meherDist}
\end{figure*}

\section{Results}

\subsection{MEHER}
\label{sec:meher}

In Figure \ref{fig:meher}, we trained RL agents across all spawn and prey policy conditions while varying the S-ratios between 0 and 1.  For all environments, at least one of the S-ratio conditions successfully learned to be nearly 100\% successful.  While the agents in the 0.5 S-ratio condition achieved an average maximum median success rate of 0.92, agents in the 0.6 S-ratio condition (the highest performing condition), achieved an average maximum median success rate of 0.99.  Agents in the 0.5 S-ratio condition did learn approximately 11\% faster than agents in the 0.6 S-ratio condition, on average.

In Figure \ref{fig:meherDist}, the results are summarized using the $M_c$ metric described in Equation \ref{eq:mc}.  When displayed like this, information theory predicts that the bars should be arranged into a parabola, with a maximum at $x = 0.5$.  Additionally, information theory suggests that S-ratios of 0 and 1 should result in $M_c = 0$.  While some conditions do display this pattern, for many of the conditions, the S-ratios greater than 0.5 perform better than S-ratios less than 0.5.  This deviation (where higher S-ratios perform better) is particularly apparent for the Spawn Random conditions.  Across all simulation conditions, the mean of the bar plots (which would correspond to the maximum of a parabola, if a parabolic fit were appropriate) was 0.60 $\pm$ 0.06 (mean $\pm$ standard deviation).  However, the prediction that $M_c = 0$ for S-ratios of 0 and 1 seems appropriate, given that only in one condition (Spawn Apart, Attract) does the maximum median reward approach 1 for S-ratios of 0 or 1.  Supplementary Figures \ref{fig:uTargetReward} \& \ref{fig:uTargetTime} show the components of $M_c$ (maximum median reward, and time to learn), respectively.

\subsection{Targeted MEHER}
\begin{figure*}
    \centering
    \includegraphics[width=\textwidth]{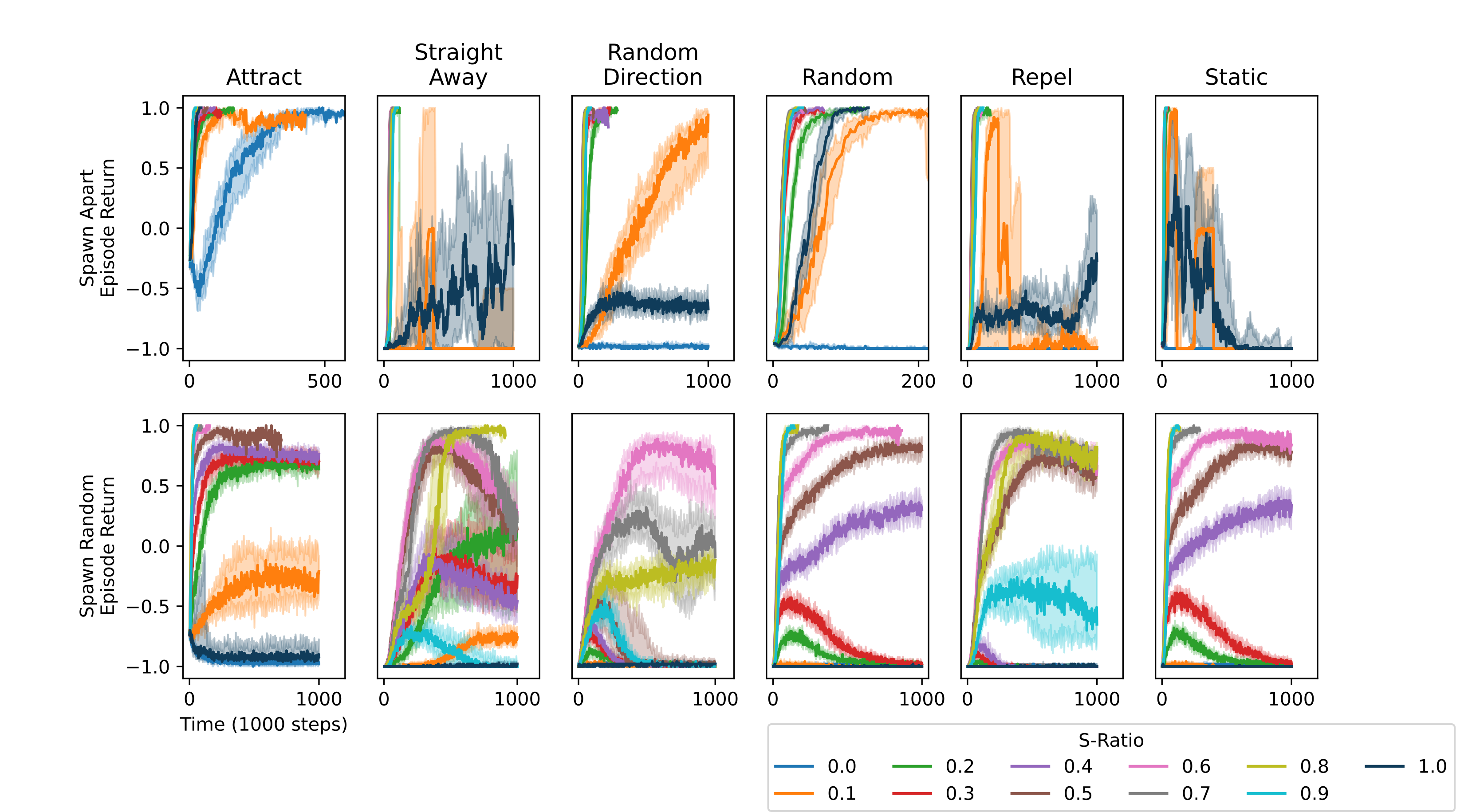}
    \caption{Learning curves for agents trained using targeted MEHER, where targets for artificial failures are generated near the target, rather than far away.  Comparing to un-targeted MEHER (Figure \ref{fig:meher}), targeted MEHER resulted in very few changes.}
    \label{fig:targetedMeher}
\end{figure*}

\begin{figure*}
    \centering
    \includegraphics[width=\textwidth]{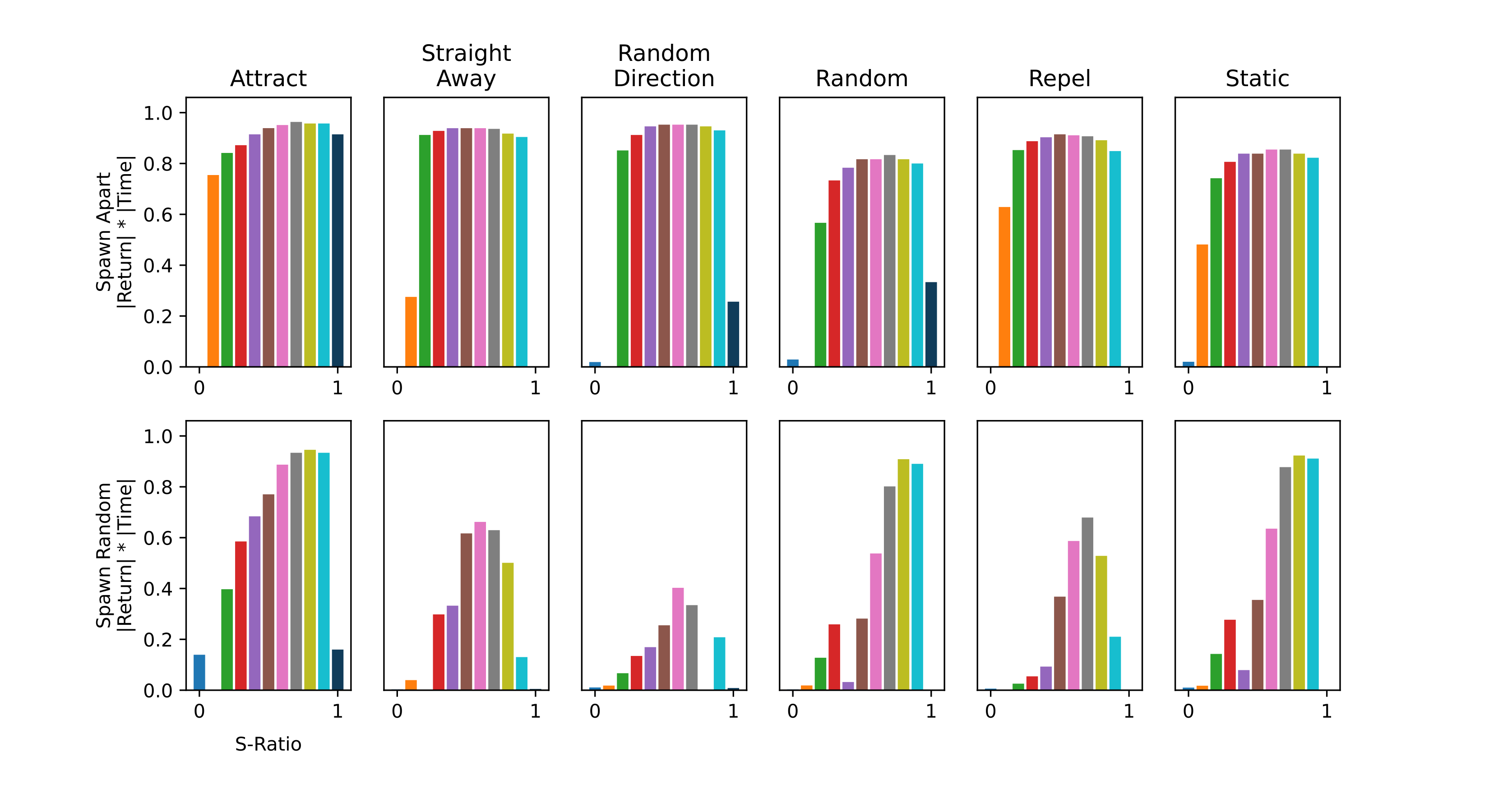}
    \caption{$M_c$ metrics for S-ratios in the range $[0, 1]$ when using targeted MEHER.  The shape of the bar plots, rather than the magnitudes of the columns, should be interpreted.  Information theory predicts a parabolic shape to the bar plot, maximized at a S-ratio of 0.5, with $M_c$ values of 0 for S-ratios of 0 and 1.  The actual shapes of the bar plots differ somewhat, and higher S-ratios tend to perform better than predicted.  The shapes of the bar plots look similar to the bar plots for un-targeted MEHER (Figure \ref{fig:meherDist}).}
    \label{fig:targetedMeherDist}
\end{figure*}

Results in Section \ref{sec:meher} suggested that successes were more informative than failures, shifting the bar plot of $M_c$ values, as a function of S-ratios, to the right.  In an attempt to rectify this departure from the information theory model, we attempted to make the failures more informative by drawing failure goals from a distribution near the desired goal.  As can be seen by comparing Figure \ref{fig:targetedMeher} to Figure \ref{fig:meher}, this type of targeted failure goal generation generally did not improve performance.

In Figure \ref{fig:targetedMeherDist}, the results are summarized using the $M_c$ metric described in Equation \ref{eq:mc}.  Compared to Figure \ref{fig:meherDist}, which shows results for the un-targeted condition, the relative heights of the bars do not change greatly.  Across all simulation conditions, the mean of the bar plots (which would correspond to the maximum of a parabola, if a parabolic fit were appropriate) is 0.59 $\pm$ 0.06 (mean $\pm$ standard deviation), which is nearly the same as the un-targeted MEHER condition. Supplementary Figures \ref{fig:TargetReward} \& \ref{fig:TargetTime} show the components of $M_c$ (maximum median reward, and time to learn), respectively, for the targeted MEHER condition.

\subsection{PPO-HER to PPO}
\begin{figure*}
    \centering
    \includegraphics[width=\textwidth]{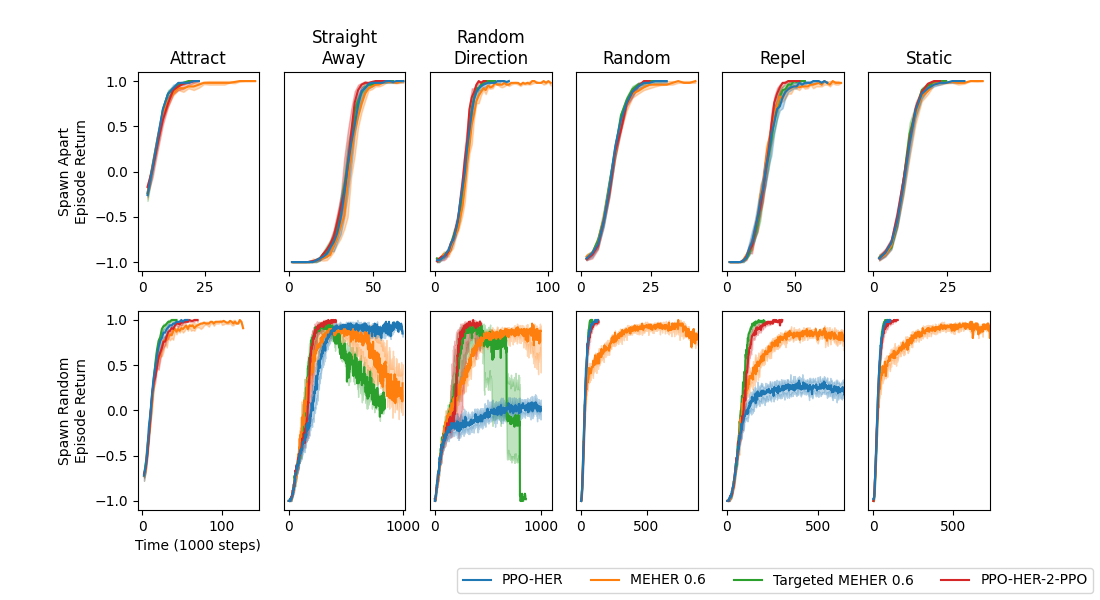}
    \caption{Learning curves for agents trained using S-ratios in the range $[0, 1]$ that were trained using PPO-HER, un-targeted maximum entropy PPO-HER, targeted maximum entropy PPO-HER, or PPO-HER-2-PPO.  PPO-HER-2-PPO performs as well as other methods. Bold lines and shaded regions represent medians and interquartile ranges, respectively.}
    \label{fig:ppoHer2Ppo}
\end{figure*}

\begin{figure*}
    \centering
    \includegraphics[width=\textwidth]{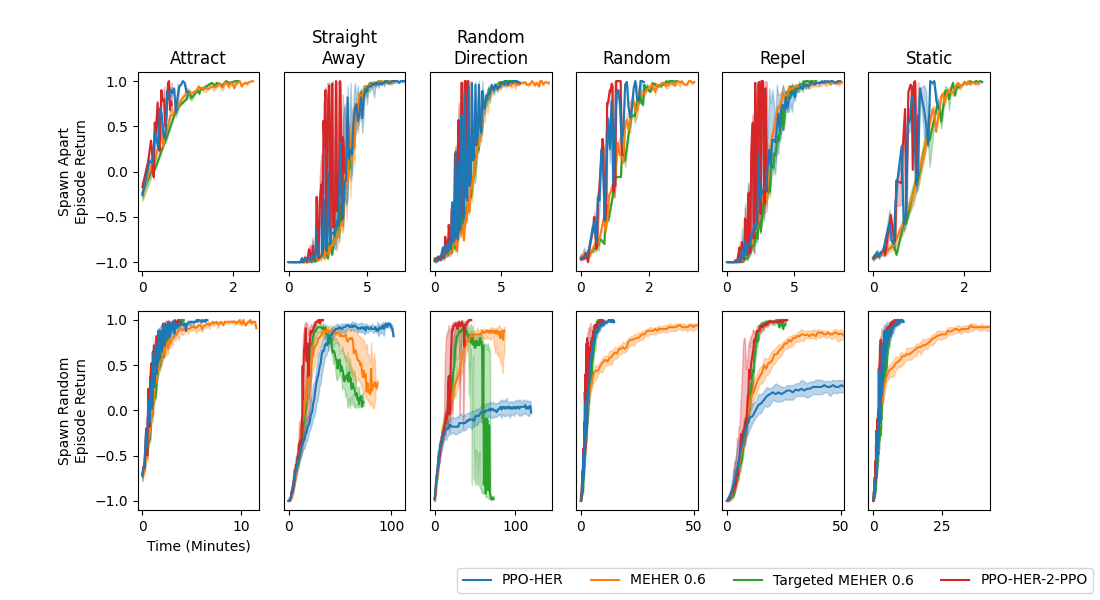}
    \caption{Learning curves for agents trained using S-ratios in the range $[0, 1]$ that were trained using PPO-HER, un-targeted maximum entropy PPO-HER, targeted maximum entropy PPO-HER, or PPO-HER-2-PPO.  PPO-HER-2-PPO performs as well as other methods.  Performance is plotted against clock time to highlight the performance of PPO-HER-2-PPO. High-frequency artifacts are caused by early stoppage of high-performing agents.  Bold lines and shaded regions represent medians and interquartile ranges, respectively.}
    \label{fig:ppoHer2PpoTime}
\end{figure*}

To make PPO-HER more clock-time efficient, we developed an alternative method to MEHER that turns HER off after the agent is successful 50\% of the time.  We compared the performance of this so-called PPO-HER-2-PPO method to un-targeted and targeted MEHER.  We chose to use the 0.6 S-ratio for the MEHER conditions since this S-ratio achieved the highest mean maximum median reward.  As shown Figure \ref{fig:ppoHer2Ppo}, PPO-HER-2-PPO performed as well as other methods, reaching the maximum median reward in 10 of the 12 conditions.  While achieving similar maximum median rewards, PPO-HER-2-PPO was able to learn in 45\% to 56\% of the clock time required by targeted or un-targeted MEHER, respectively, as shown in Figure \ref{fig:ppoHer2PpoTime}.

\section{Discussion}

\subsection{MEHER}
Using information theoretic entropy, we theorized that we could improve HER (when combined with PPO) by modifying the percentage of successes that existed in the buffer.  Specifically, our theoretical analysis predicted 3 things:

\begin{enumerate}
    \item Plots of maximum median rewards, time to learn, and $M_c$ would be approximately parabolic when plotted as functions of the S-ratio
    \item The maximum of the parabola (the highest performing condition) would occur when the S-ratio was 0.5 because an S-ratio theoretically maximizes the amount of information in the reward signal
    \item When the S-ratio was either 0 or 1, the entropy of the reward signal would be 0, resulting no learning (x-intercepts of the parabolas)
\end{enumerate}

This parabolic models seems to be reasonably appropriate for many of the environmental conditions, with some notable deviations.  Firstly, while the bar plots for the Spawn Apart conditions are approximately parabolic and centered at 0.5, the bar plots for Spawn Random conditions tend to be shifted to the right, suggesting that successes are more informative than failures.  Secondly, for a minority of conditions (most notably, the Attract Spawn Apart environment), the 1 S-ratio condition performs much better than the predicted success rate of 0 - a pattern that also suggests that successes are generally more informative than failures.

It may be possible to explain these discrepancies by considering that, in interesting RL environments, random walks rarely result in successes, meaning that the prior probability of being successful is very small.  Consider, for instance, an RL agent that is initialized with a random policy that is rarely successful.  Now, break history into 2 epochs, $t_T$, after training begins, and $t_I$, where the agent policy is evaluated, but the agent is not trained.  If $t_I >> t_T$, the frequentist probability of observing a success, $p(s)$ remains close to the the probability of observing a success, given that we are in the Initialize (not Training) phase $p(s|I)$.  Using this interpretation, $p(s)$ is always low, meaning that the observation of a success is always informative compared to the observation of a failure.  This explanation is supported by the fact that the Spawn Random environments are empirically more difficult to solve than the Spawn Apart environments, and the Spawn Random environments benefit more from higher S-ratios.

This theory suggests that HER is suboptimal because it adds transitions with resampled goals to the buffer in an un-principled way.  In the original paper \cite{andrychowicz2017hindsight}, the performance of HER was sensitive to 2 main hyperparameters: the method by which goals are resampled and the the number of resampled transitions per original transition in the buffer.  Tuning of these hyperparameters results in good performance in many scenarios.  While the original paper suggests hyperparameters that work well across many different scenarios, our own work suggests that these hyperparmeters may need to be adjusted, for instance, when transitioning from off-policy to on-policy algorithms.  By using the principle of maximum entropy, we can decrease the number of hyperparameters to just 1 (the S-ratio), and we can provide a single value (S-ratio = 0.6) that works well across a suite of Predator-Prey environments.  
By using MEHER, we provide a more principled way to choose a single hyperparameter that can be theoretically justified.

Our methods bear some resemblance to the methods used in \cite{dai2020episodic}, although we used a pure RL methodology, whereas Dai et al. uses an imitation learning component.

\subsection{Targeted MEHER}
To test if information theory can explain the behavior of the agents in the Spawn Random environments, which tend to perform better with higher S-ratios, we attempted to make failures more informative by choosing resampled failure goals from locations near the achieved goal.  The rationale was that it is easier to learn from a near miss than a far miss.  However, targeted MEHER resulted in almost no change to the performance of the agents.  Despite this apparent failure, this result supports the information theoretic interpretation of HER since resampled failure goals, regardless of their positions in the environment, do not change the relative probabilities of observing symbols (success or failure) in the reward signal.

\subsection{PPO-HER-2-PPO}

In general, HER works by increasing the number of successes in the buffer.  This is particularly important at the beginning of training, since relatively few successes will be achieved by chance, and successes generally contain a lot of information.  However, as training progresses, HER may continue to increase the ratio of successes in the training buffer, which may result in sub-optimal learning.  While MEHER is generally successful at promoting learning, HER requires additional computation.  This overhead is likely relatively high for PPO because PPO requires us to calculate the probability of an action, given the current state, and when the goal is changed, the state is changed.  Because HER may actually hurt performance in the last half of training, and because it adds computational overhead, we explored the possibility of improving performance by simply turning HER off in the last half of training (once a success rate of 0.5 was achieved).

As shown in Figure \ref{fig:ppoHer2Ppo}, this so-called PPO-HER-2-PPO method works as well as MEHER in terms of achieving maximum rewards.  And, as shown in Figure \ref{fig:ppoHer2PpoTime}, PPO-HER-2-PPO can achieve these results in significantly less clock time, especially when the environment simulation is quick, as is the case for this study.

We suspect that PPO-HER-2-PPO out-performed MEHER, despite the information theoretic optimally of MEHER, because of implementation details.  
To achieve the correct S-ratio, we resampled goals for each episode, and then we removed transitions from the buffer.  By removing transitions, we sometimes discarded actual transitions instead of HER transitions.  Note that, when using the ``final'' HER method (as we did), resampled goals for successful trials are the same as the original goals, resulting in a copy of the transition being added to the buffer.  While transitions were randomly removed from the buffer, it is likely that some buffers contained samples that were predominantly from a few (successful) episodes.  This would decrease the diversity of observations and actions in the buffer, which, from an information theoretic perspective, is sub-optimal, since we want actions and observations to be drawn from a uniform distribution (i.e., we want diversity in our observations and actions).   Secondly, by adding multiple instances of the same transition to the buffer, this could promote overtraining on those transitions, particularly for PPO buffers, which are limited in size.  PPO-HER-2-PPO may avoid these issues by simply turning off HER.

\subsection{Limitations and Future Work}
The purpose of this work was to explore how we could optimize HER for accelerating PPO since our previous studies \cite{crowder2024ppoher} yielded the surprising result that HER could, in fact, accelerate PPO, despite the fact that HER violates the on-policy assumption of PPO.  Future work should replicate these results with other RL algorithms, including off-policy algorithms such as SAC.  These results should also be repeated in more complex, standardized environments such as the Fetch robotic arm environments \cite{plappert2018multi}, especially because our previous work suggested that PPO-HER was ineffective at Fetch.

In this work, we were able to use HER to manipulate the information contained within the reward signal.  In addition to the reward signal, the neural networks underlying deep RL algorithms, including PPO, use the action and observation signals to learn.  We were unable to explore how maximum entropy, as applied to the action vector, would affect results.  And, while the Spawn Random positions have more entropy than the Spawn Apart positions, the Spawn Random environments require the RL agent to generalize across more initial states, which makes the task more difficult, and adds a confounding variable that prevents us from directly comparing the 2 spawners to draw conclusions about maximum entropy in the observation signal.  Future studies should consider how to investigate maximum entropy within the action and observation signals.  Offline RL may be useful for this purpose, although careful experimental design will be required to dissociate entropy and diversity of states.

To derive our information theoretic explanation for the performance of MEHER, we simplified the reward signal by considering it to be a return signal that only contained values of $\{-1, 1\}$.  Because of temporal difference learning, it is unclear if this simplification is valid for the environments used here.  For instance, if two successful episode last for 2 steps and 3 steps, with a discount factor of $\gamma=0.90$, the time-discounted reward signals will be $\{1, 0.9\}$ and $\{1, 0.9, 0.81\}$, respectively.  This means that reward signals for long episodes will contain rarer symbols than reward signals for shorter episodes, which would imply that there is more information in long episode reward signals.  But, this would seem to contradict our present finding that, particularly for more difficult environments, successful episodes (which likely terminate sooner) contain more information.  It should be possible to explore the importance of our episode return assumption by creating spawners where all targets can be reached in a minimum of $n$ steps and where the timeout (which ends the episode) is also $n$ steps, meaning that every episode is the same length.  In this case, the time-discounted rewards will all contain the same number of symbols, all appearing with equal probability.  Information theory assumes that the signal is a random variable, and because time-discounted rewards will appear in descending order (reading the signals backwards in time), the vector is not truly random.  However, if we condense each episode into a single symbol, represented by the terminal rewards (i.e., the episode return) of $\{-1, 1\}$, then the signal will be random.  This collapse can be made without making any assumptions, but only if each episode is the same length.

One of the most intriguing results from this work was the fact that PPO-HER-2-PPO provides better performance than MEHER (targeted or un-targeted) as well as standard PPO-HER.  For this limited study, we turned HER off once the agent was able to achieve success 50\% of the time, which was based on our information theory interpretation of HER.  However, empirical results suggest that a S-ratio of 0.6 may be more efficacious across all tested experimental conditions.  Future experiments should titrate the success rate threshold at which HER is turned off, which could provide additional evidence to support or refute our present finding that successes generally contain more information than failures.

Lastly, while we investigated maximum entropy as it pertains to HER, our findings should be explored more generally in the fields of RL and machine learning.  Existing work to this end includes maximum entropy RL, which attempts to maximize the entropy within the policy \cite{ziebart2008maximum}, which can make RL policies more robust to dynamic environments \cite{eysenbach2021maximum}.  Even in the initial HER paper, there are some results demonstrating that a diversity of initial states can allow HER to converge more quickly \cite{andrychowicz2017hindsight}.  We hope that these present results, and the accompanying theory, will provide a framework to better interpret previous results, stimulate new experiments, and contribute to the development of more sample-efficient RL algorithms.

\section{Conclusions}
We used information theory to derive a new type of HER, called MEHER that improves the performance of PPO-HER in a Predator-Prey environment.  We provided evidence, and a theoretical explanation, for the fact that successes in RL contain more information than failures.  And, we created a new algorithm, called PPO-HER-2-PPO, that performs better than MEHER while also being more clock time efficient.

\bibliographystyle{IEEEtran}
\bibliography{bib}

\appendix

\begin{figure*}
    \centering
    \includegraphics[width=\textwidth]{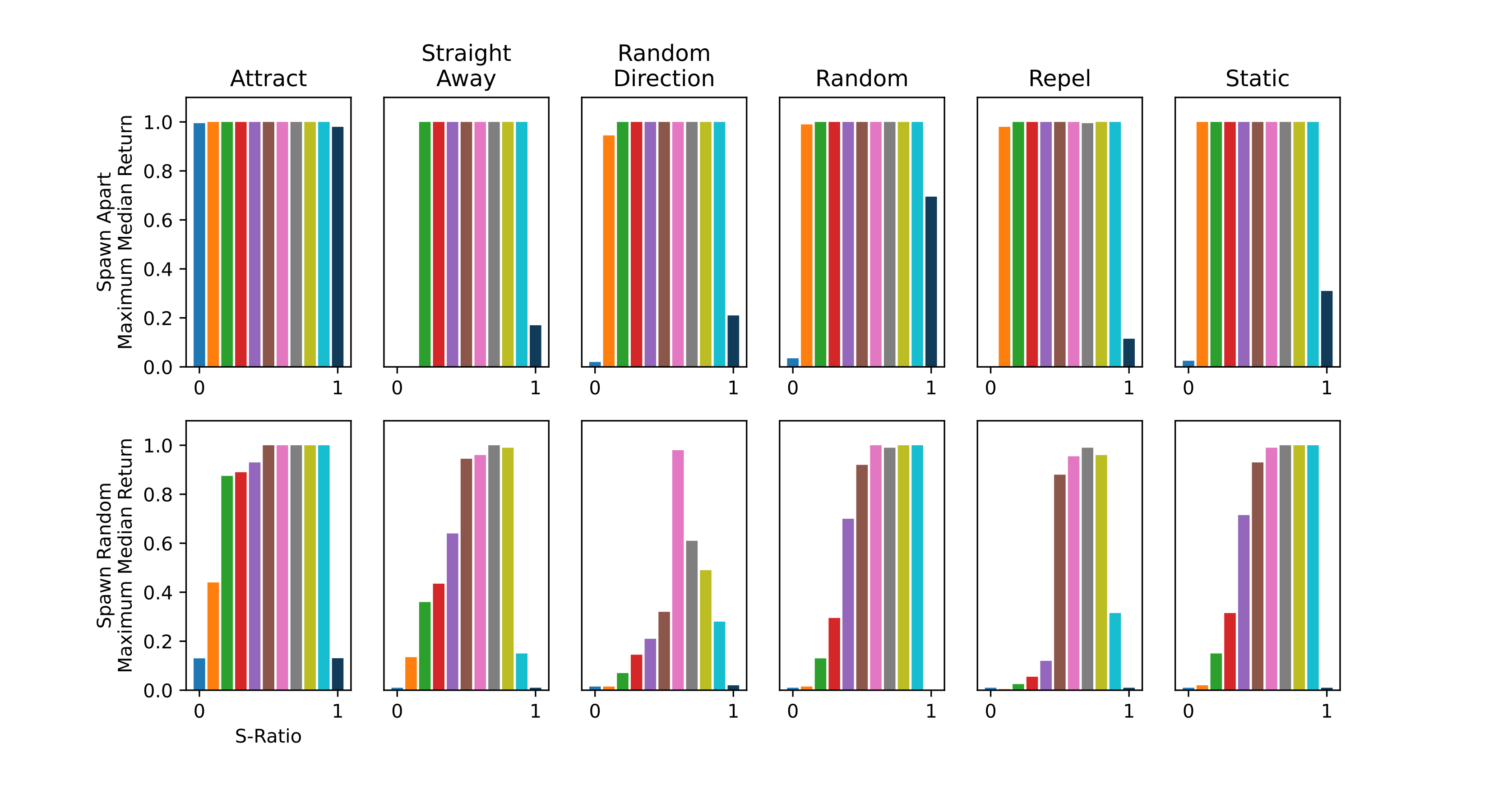}
    \caption{Maximum median reward as a function of S-ratio for un-targeted MEHER.}
    \label{fig:uTargetReward}
\end{figure*}

\begin{figure*}
    \centering
    \includegraphics[width=\textwidth]{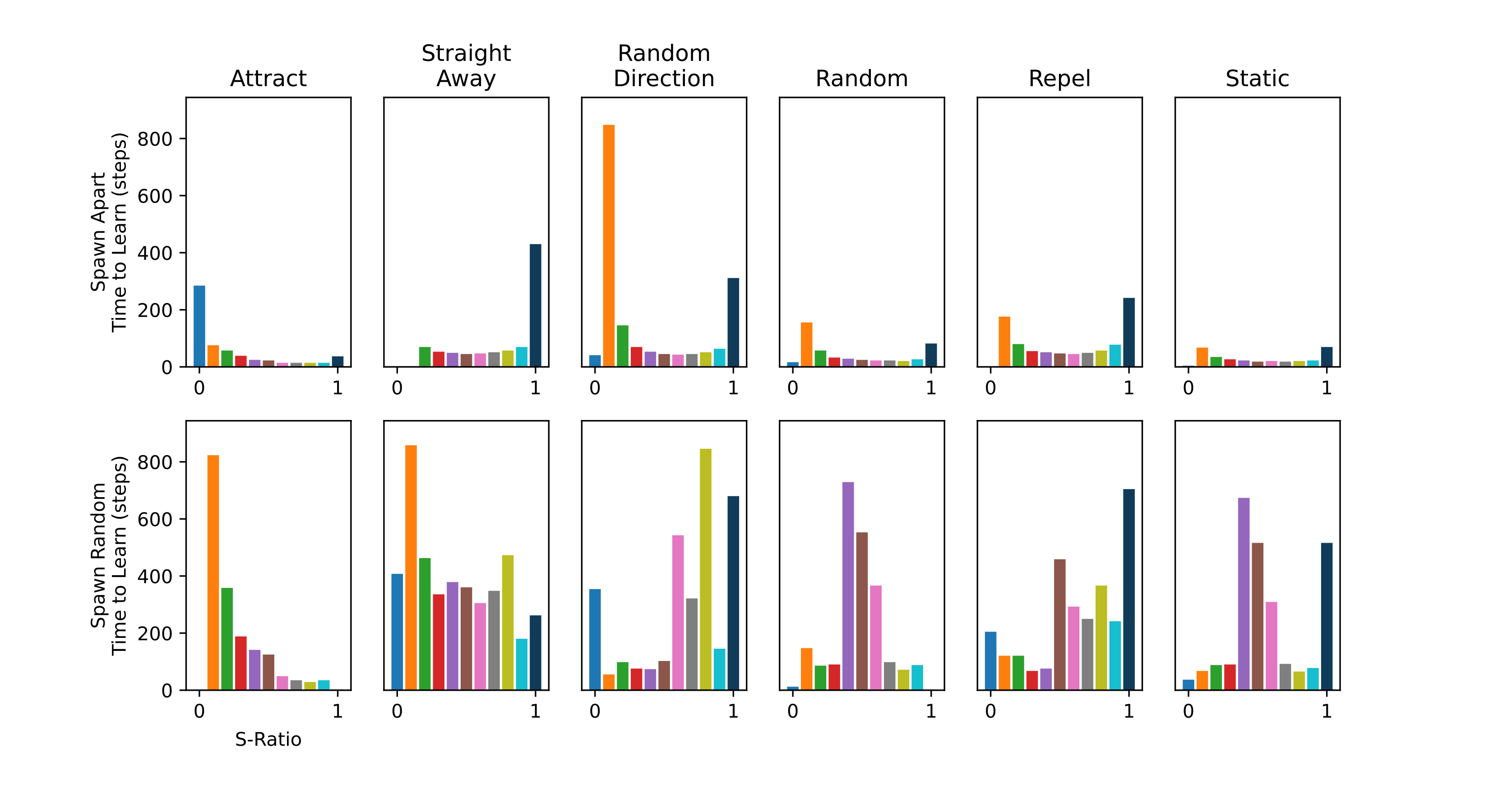}
    \caption{Time to learn as a function of S-ratio for un-targeted MEHER.}
    \label{fig:uTargetTime}
\end{figure*}

\begin{figure*}
    \centering
    \includegraphics[width=\textwidth]{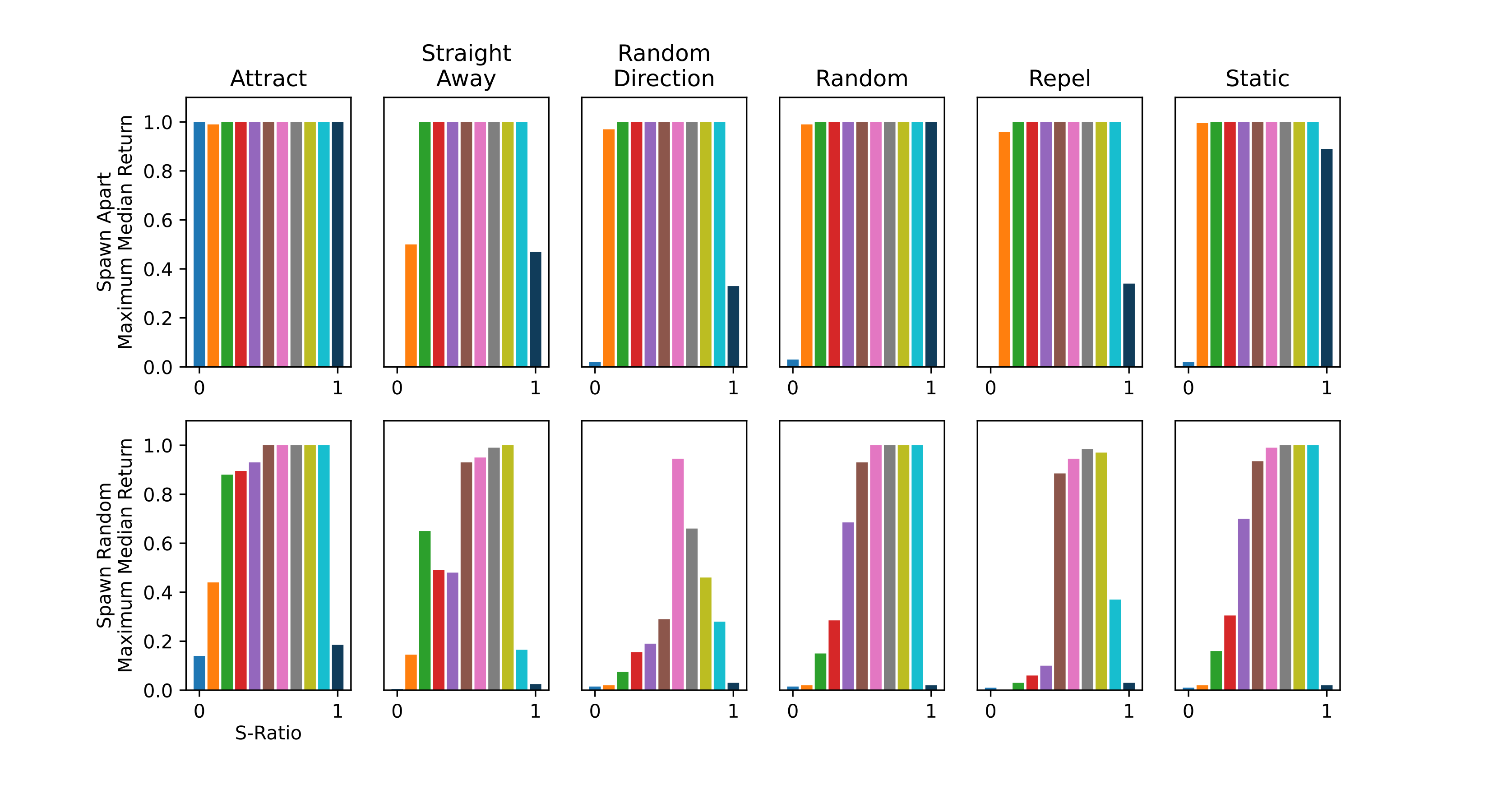}
    \caption{Maximum median reward as a function of S-ratio for targeted MEHER.}
    \label{fig:TargetReward}
\end{figure*}

\begin{figure*}
    \centering
    \includegraphics[width=\textwidth]{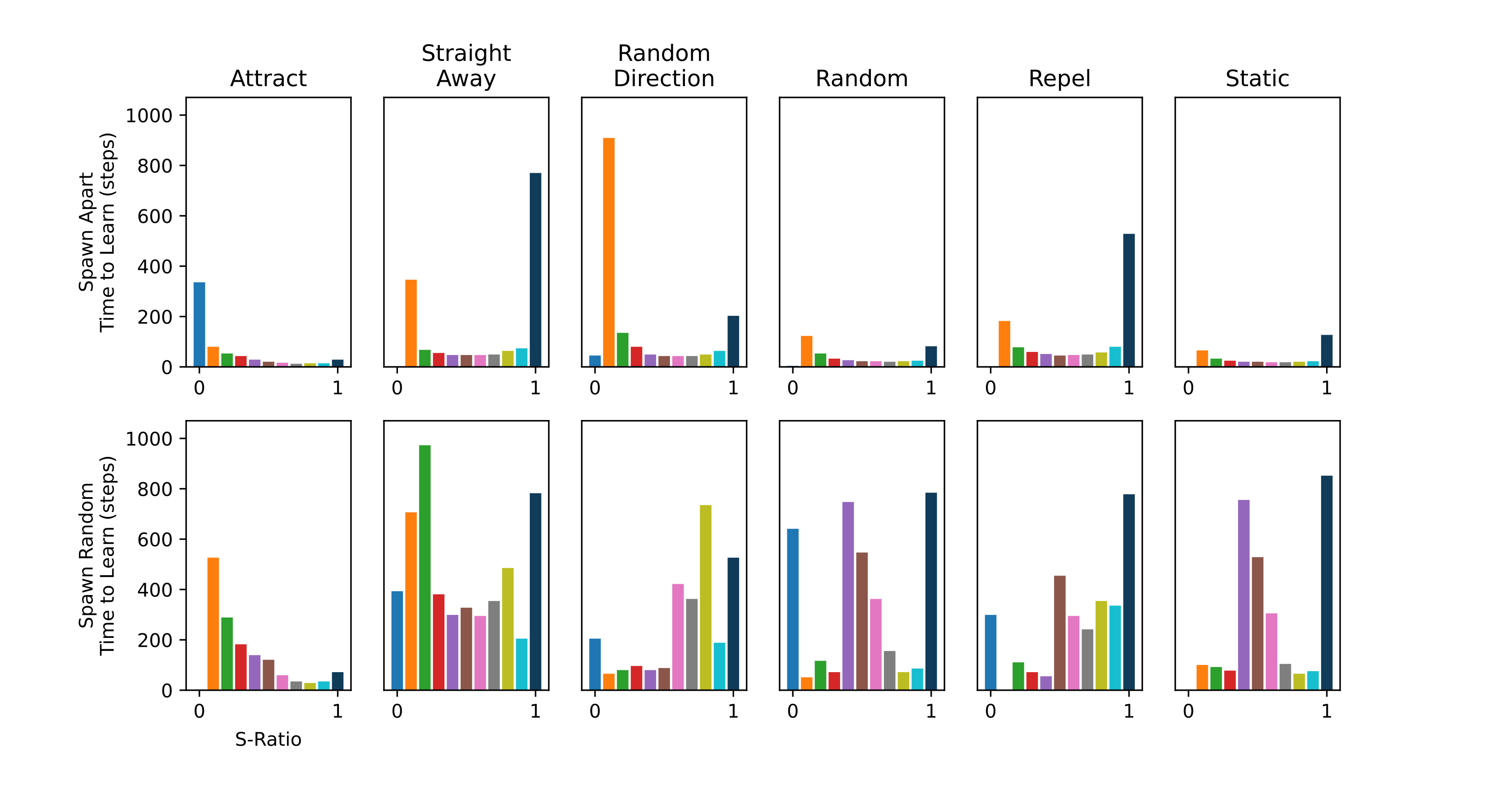}
    \caption{Time to learn as a function of S-ratio for targeted MEHER.}
    \label{fig:TargetTime}
\end{figure*}

\end{document}